\def\year{2022}\relax
\documentclass[letterpaper]{article} 
\usepackage{aaai22}  
\usepackage{times}  
\usepackage{verbatim}
\usepackage{helvet}  
\usepackage{courier}  
\usepackage[hyphens]{url}  
\usepackage{graphicx} 
\usepackage{natbib}  
\usepackage{caption} 
\DeclareCaptionStyle{ruled}{labelfont=normalfont,labelsep=colon,strut=off} 
\usepackage{algorithm}
\usepackage{algorithmic}
\usepackage{xcolor}
\usepackage{amsmath}
\usepackage{amssymb}
\usepackage{booktabs}
\usepackage{newfloat}
\usepackage{listings}
\floatstyle{ruled}
\newfloat{listing}{tb}{lst}{}
\floatname{listing}{Listing}
\title{Role of Human-AI Interaction in Selective Prediction}
\author{
    Elizabeth Bondi, \textsuperscript{\rm 1\footnote{Work done at DeepMind}} Raphael Koster, \textsuperscript{\rm 2} Hannah Sheahan, \textsuperscript{\rm 2} Martin Chadwick, \textsuperscript{\rm 2} Yoram Bachrach, \textsuperscript{\rm 2} Taylan Cemgil, \textsuperscript{\rm 2} Ulrich Paquet, \textsuperscript{\rm 2} Krishnamurthy (Dj) Dvijotham \textsuperscript{\rm 3\footnotemark[1]}
}
\affiliations{
\textsuperscript{\rm 1}Harvard University\\
\textsuperscript{\rm 2}DeepMind\\
\textsuperscript{\rm 3}Google Brain \\
ebondi@g.harvard.edu, \{rkoster, hsheahan, mjchadwick, yorambac, taylancemgil, upaq\}@deepmind.com, dvij@google.com
}

\newcommand{\serengeti}{Serengeti} 

\newcommand{\nomsg}{NM}
\newcommand{\both}{BM}
\newcommand{\deferonly}{DO}
\newcommand{\predonly}{PO}
\newcommand{\msg}{SPM}
\newcommand{\defermain}{deferral status}
\newcommand{\predmain}{prediction}

\begin{document}

\maketitle

\begin{abstract}

Recent work has shown the potential benefit of  selective prediction systems that can learn to defer to a human when the predictions of the AI are unreliable, particularly to improve the reliability of AI systems in high-stakes applications like healthcare or conservation. However, most prior work assumes that human behavior remains unchanged when they solve a prediction task as part of a human-AI team as opposed to by themselves. We show that this is not the case by performing experiments to quantify human-AI interaction in the context of selective prediction. In particular, we study the impact of communicating different types of information to humans about the AI system's decision to defer. Using real-world conservation data and a selective prediction system that improves expected accuracy over that of the human or AI system working individually, we show that this messaging has a significant impact on the accuracy of human judgements. Our results study two components of the messaging strategy: 1) Whether humans are informed about the prediction of the AI system and 2) Whether they are informed about the decision of the selective prediction system to defer. By manipulating these messaging components, we show that it is possible to significantly boost human performance by informing the human of the decision to defer, but not revealing the  prediction of the AI. We therefore show that it is vital to consider how the decision to defer is communicated to a human when designing selective prediction systems, and that the composite accuracy of a human-AI team must be carefully evaluated using a human-in-the-loop framework.

\end{abstract}
\begin{figure*}
    \centering
    \includegraphics[width=0.98\textwidth]{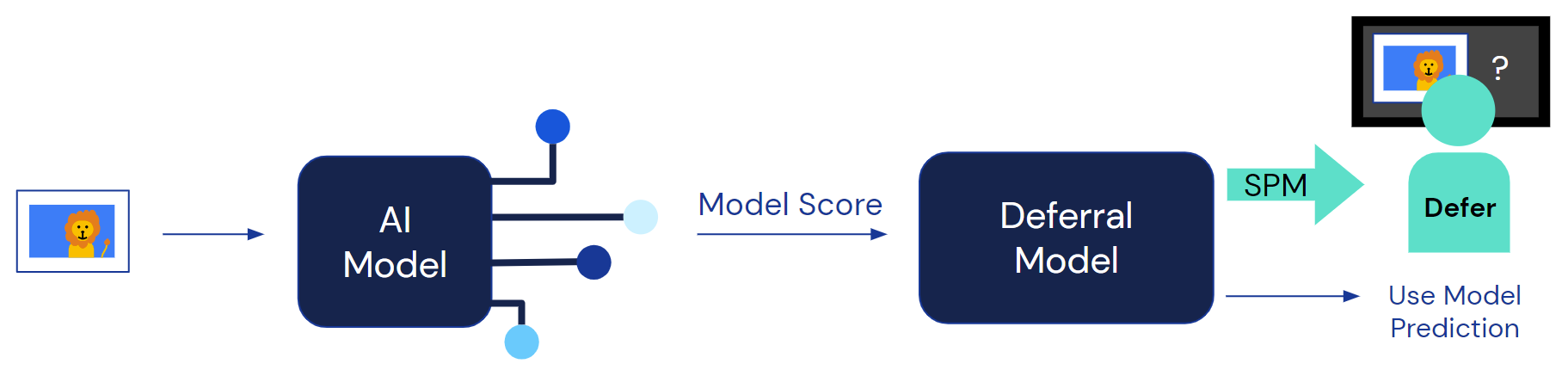}
    
    \caption{Deferral workflow: (1) obtain AI model prediction for given input; (2) use model score as input to deferral model to decide whether to defer; (3) defer to humans if necessary, using \msg~based on the \predmain~and \defermain.}
    \label{fig:workflow}
\end{figure*}
\section{Introduction} 


Despite significant progress in machine learning-based AI systems, applications of AI to high stakes domains remains challenging. One of the main challenges is in assessing the reliability of predictions made by a trained machine learning system, particularly when there is a distribution shift between the data the system was trained with and data encountered at deployment. In such situations, communicating the uncertainty associated with ML predictions appropriately is critical \citep{kompa2021second, grote2021trustworthy}.

Given the difficulty of communicating probabilities to human users \citep{galesic2010statistical,zhang2012ubiquitous}, a pragmatic alternative is to determine whether an AI system is more likely to make an erroneous prediction than a human, and defer to a human in such cases. A number of such settings have been studied in the literature, including selective prediction \citep{wiener2013theoretical}, learning to defer \citep{mozannar2020consistent} and classification with a reject option \citep{chow1970optimum}. While there are nuances that differentiate these works, in this paper, we will collectively refer to this body of literature as selective prediction and only emphasize the differences where relevant to our work. A related line of work considers human-AI teams  in which humans receive AI assistance but make the final decision  \citep{bansal2021most}, and systems in which the AI makes the final decision with human input \citep{wilder2020learning}. These prior works either: a) Assume that the human behaves identically even when they know that they are part of a human-AI team or b) Assume a utility-maximizing model for a human decision maker. 

However, it has been documented that human-AI interaction may be more complex due to a range of cognitive phenomena. For example, humans have been shown to rely excessively on AI predictions (anchoring bias) \citep{rastogi2020deciding,logg2019algorithm}, or even distrust AI predictions after observing AI mistakes \citep{dietvorst2015algorithm}. Some work has begun to investigate solutions to these issues \citep{buccinca2021trust,rastogi2020deciding}, however they have not focused on selective prediction systems, and context is critical. 

In this work, we focus on binary classification tasks and study selective prediction systems (Fig. \ref{fig:workflow}) that determine whether to rely on the outputs of an AI or defer to a human. To evaluate the overall performance of such a selective prediction system, it is important to model how the messaging, hereafter referred to as selective prediction messaging (\msg), that communicates the decision to defer impacts human accuracy. We run experiments with human subjects solving a challenging binary prediction task (that of detecting whether an animal is present in a camera trap image) and study the impact of different choices for communicating the deferring AI system's decision. We then perform statistical analysis on the human responses under various choices of \msg, and show that the choice of \msg~significantly impacts human performance.  Our results isolate two ingredients for a statistically effective communication strategy, that is, communicating that an AI system deferred (\defermain) and the AI system's predictions. Manipulating these leads to a boost in overall accuracy. We provide some plausible explanations for this phenomenon and suggest avenues for further work.
Our contributions are therefore as follows:
\begin{itemize}
   \item We develop  and  implement a balanced experimental  design  that  can be  used  to  measure  the  impact  of  \msg~on the accuracy of selective prediction systems.
   \item We investigate the consequences of \msg~on joint human-AI performance in a conservation prediction task, as opposed to prior work which assumes human behavior remains static during deferral.
   \item We discover two separable \msg~ingredients, \defermain~and \predmain, which have distinct, significant effects on human performance, and demonstrate that manipulating these ingredients leads to improved human classifications in a human-AI team, implying that the setup and the information given to humans during such tasks has a large  impact on the performance of a human-AI team.
   \item We suggest that our results may relate to a more general property of naturalistic datasets, that in conditions that are ambiguous, 
   sharing AI model predictions with humans can be detrimental.

\end{itemize}

\section{Related Work} 
We start by describing prior work considering different potential roles in human-AI teams, namely, decision aids and systems in which an AI model defers to a human on challenging cases only. We then discuss supporting human-AI decision-making and selective prediction algorithms.

\paragraph{Decision Aids:} 
Potential deployment scenarios for human-AI teams that have been discussed in the literature vary greatly depending on the application. 
One increasingly common scenario, particularly in high-risk domains, is that of AI systems serving as decision aids to humans making a final decision. For example, \citet{green2019principles} explore the scenario of human decision-makers using a risk assessment model as a decision aid in financial lending and criminal justice (specifically, pretrial detention) settings. They found that humans were biased and that they failed to properly evaluate or take model performance into account, across different messaging conditions communicating the model's predicted risk. A decision aid for risk assessment is similarly studied in \citet{de2020case}, particularly in the real-world domain of child maltreatment hotline screening. Humans indeed changed their behavior based on the risk assessment tool, but they were able to identify model mistakes in many cases. 
\citet{gaube2021ai} conduct experiments to measure the interaction between medical AI systems and clinicians. Radiologists who participated reported AI advice as lower quality than human advice, though all advice truly came from humans. Furthermore, clinician diagnostic accuracy was reduced when they were given incorrect predictions from the AI system. 
Even with humans making the final decision in each case, there is a great deal of variability between these AI systems, their impacts, and their application domains.

\paragraph{Deferral:} 
AI systems have also played a slightly different role by making decisions in straightforward cases and deferring to human decision makers otherwise, which is most similar to our scenario. \citet{wilder2020learning} defer to an expert on cases that are best suited for human decision-making compared to model-based decision-making (determined using end-to-end learning), yet the final system is evaluated on historical human data, meaning humans did not know they were part of a human-AI team while labeling.  \citet{keswani2021towards} propose deferral to multiple experts, including a classifier, by learning about the experts from their decisions.  While Amazon Mechanical Turk was used to collect labels from participants to train and evaluate this model, it is similar to using historical human labels, as the humans again completed the task without knowing the deferral status or model prediction. Such deferral models are not yet widely deployed, nor have their impacts on humans been studied. 

\paragraph{Supporting Human-AI Decision-Making:}
To mitigate some of the known negative impacts of AI on human decision-making, it may be beneficial to present humans with further information, such as uncertainty. 
\citet{bhatt2021uncertainty} find that humans may unreliably interpret uncertainty estimates, but that the estimates may increase transparency and thereby performance. In a system used by people without a background in statistics, for example, presenting categories (such as \defermain) may make it easier to interpret.  
\citet{bhatt2021uncertainty} advocate for testing this with humans, including of different skill levels and in different domains. 
Further suggestions for positive human-AI interaction are presented in \citet{amershi2019guidelines}, including to ``Show contextually relevant information,'' and ``Scope services when in doubt.'' We are interested in finding the best way to implement these ideas with \msg.

\paragraph{Selective Prediction:} 
Selective prediction can be traced back to the seminal work of \citet{chow1970optimum}, where theoretical properties of classifiers that are allowed to ``reject'' (refrain from making a prediction) and ideal rejection strategies for simple classifiers are investigated. In these settings, the main metrics are the accuracy of the classifier on the non-rejected inputs and the rate of rejection, and the natural trade-off between these. A recent survey of theoretical work in this area can be found in \citet{wiener2013theoretical}. We omit an extensive review of selective prediction literature, but acknowledge there is additional work in this area.

More relevant to our work is the work on learning to defer \cite{madras2017predict,mozannar2020consistent}. \citet{madras2017predict} propose to defer to a human decision maker selectively in order to improve accuracy and fairness of a base classifier. \citet{mozannar2020consistent} develop a statistically consistent loss function to learn a model that both predicts and defers. 
\citet{geifman2017selective} develop deferral strategies purely based on the confidence estimate of an underlying classifier.

None of these works model the impact of deferral (or its communication to a human decision maker) on the accuracy of a human decision maker. Since this is the primary object of study in our work, we do not include a full selective prediction literature review. We use a simple confidence-based deferral strategy inspired by the work of \citet{geifman2017selective}, but our experimental design is compatible with any selective prediction system. 

\section{Background}\label{sec:background}
The primary goal of our work is to evaluate the impact
of \msg~on human performance in selective prediction systems. We design an experiment to evaluate this impact and base both our design and the questions we study on the psychological literature on joint decision-making.

\paragraph{Psychology Literature on Human Decision-Making:} The psychology literature has extensive studies on decision-making in human teams. Three psychological phenomena stand out as being relevant to our work: 1) Humans are sensitive to the specific way that a task is framed \citep{tversky1981framing}, 2) Humans are capable of flexibly deploying greater attentional resources in response to changing task demands and motivation \citep{kool2018mentallabour}, and 3) When deciding how best to integrate the decisions of others, humans take into account their own decision confidence as well as the inferred or stated confidence of other decision makers \cite{BOORMAN20131558, bahrami2010optimally}.

This suggests that \msg~
may impact how humans perceive their task, how much they trust the AI system, and ultimately how accurate their final prediction is, which is directly related to the composite performance of a selective prediction system. We consequently propose an  experimental design where we take four natural choices for  \msg~and estimate their impact on the accuracy of human labelers. 

\begin{figure}
    \centering
    \includegraphics[width=0.3\textwidth]{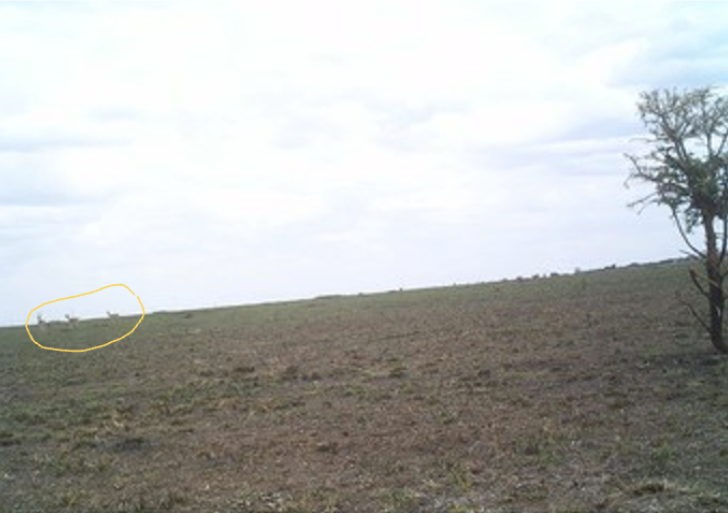}
    \caption{Example camera trap image with distant (circled) animals.}
    \label{fig:ctrap}
\end{figure}

\paragraph{Dataset:}
The dataset we use in this work is composed of images from camera traps, which are cameras triggered to capture images when there is nearby motion. These can be used to capture images of animals to understand animal population characteristics and even animal behavior, both of which are useful for conservation planning purposes.
The volume of images generated in this manner is too high for manual inspection by rangers and scientists directly involved in conservation and monitoring efforts. 

To alleviate this burden, the Snapshot Serengeti\footnote{\url{https://www.zooniverse.org/projects/zooniverse/snapshot-serengeti}} project was set up to allow volunteers to apply 
rich labels to camera trap images\footnote{\url{https://lila.science/datasets/snapshot-serengeti}}.
These labels are publicly available\footnote{ \url{https://lilablobssc.blob.core.windows.net/snapshotserengeti-v-2-0/zooniverse_metadata/raw_data_for_dryad.csv.zip}}, and ground truth comes from label consensus from multiple individuals \cite{swanson2015snapshot}.
Given these labels, AI models have been developed that automatically classify and/or detect animals in camera trap images \cite{NorouzzadehE5716,tabak2019machine,beery2020context}.

Whether relying on an AI model or volunteers, this processing is still difficult. Roughly seven out of ten images contain no animals, as they are the result of false triggers, e.g., due to heat and/or wind. 
However, it can be challenging to determine which images contain an animal at all, let alone the species, because of challenges like animal camouflage, distance to the camera, or partial visibility in the camera's field of view.  
An example camera trap image with animals on the horizon is shown in Fig. \ref{fig:ctrap}.

We consider a binary task where the bulk of blank images are first removed before images are uploaded to be labeled by volunteers or a species-identifying AI model.  
We investigate the use of a selective prediction model to filter out blank images, while prioritizing images for human review for which the blank/animal AI model is uncertain. This is similar to \citet{willi2019identifying,norouzzadeh2021deep}, as they also involve human-AI teams and remove blank images before species identification. However, 1) The human-AI teams differ, e.g., \citet{norouzzadeh2021deep} does not involve humans to remove blank images, but uses active learning for species identification, and 2) They do not focus on human behavior. 
To begin our workflow (Fig. \ref{fig:workflow}), we obtain model scores from an ensemble of AI models that filters out blank images, then develop a deferral mechanism.

\paragraph{Deferral Mechanism:} We use a selective prediction algorithm which finds the optimal threshold(s) for AI model scores to defer, as illustrated in Fig. \ref{fig:workflow}. 
Concretely, consider a binary classification task with inputs $x$ and labels $y \in \{0, 1\}$. We assume that we are given a pretrained ensemble of AI models $m$ and have access to the predictions made by a human $h$ as well. Given inputs $x$ (for example, pixels of an image), we obtain a continuous score $m(x) \in [0, 1]$ that represents the confidence of the ensemble that the label corresponding to $x$ is $1$.

The deferral mechanism we use is a simple rule-based system that identifies ranges of the model score where the model is less likely to be accurate than a human and defers on these. In particular, we use a deferral model that identifies one continuous interval in the model score space to defer on.

The deferral model is parameterized by two real numbers $0 
\leq \theta_1 \leq \theta_2$ and is defined as
\[\text{defer}(x;\theta) = \begin{cases}
1 \text{ if } m(x) \in [\theta_1, \theta_2] \\
0 \text{ otherwise }
\end{cases}\]
$\text{defer}(x)=1$ represents the decision to defer on input $x$ and $\text{defer}(x)=0$ represents the decision to predict. Given $\theta$, the AI model prediction $x$ and the prediction made by a human $h(x)$, the selective prediction system is given by
\[\text{sp}(x;\theta) = \begin{cases} h(x) \text{ if } \text{defer}(x;\theta) \\
m(x) \text{ otherwise} \end{cases}\]

Given a dataset $\mathcal{D}$ of inputs, corresponding model scores, ground truth labels, and human labels, we choose $\theta$ by solving the following optimization problem:
\[\max_{\theta} \text{Accuracy}(\mathcal{D};\theta) \text{ subject to } \text{DeferralRate}(\mathcal{D};\theta) \leq r\]
where $\text{Accuracy}(\mathcal{D};\theta)$ refers to the accuracy of the selective prediction system $\text{sp}$ with parameters $\theta$ on the dataset, $\text{DeferralRate}(\mathcal{D};\theta)$ refers to the fraction of points in the dataset for which $\text{defer}(x;\theta)=1$,  and $r$ is a bound on the deferral rate, reflecting the acceptable level of human effort or budget constraints on hiring human decision makers.

This optimization problem can be solved in a brute force manner by considering a discrete grid on the $[0, 1]$ interval and going over all possible choices for the two thresholds $\theta$.

\begin{figure}
    \centering
    \includegraphics[width=0.5\textwidth]{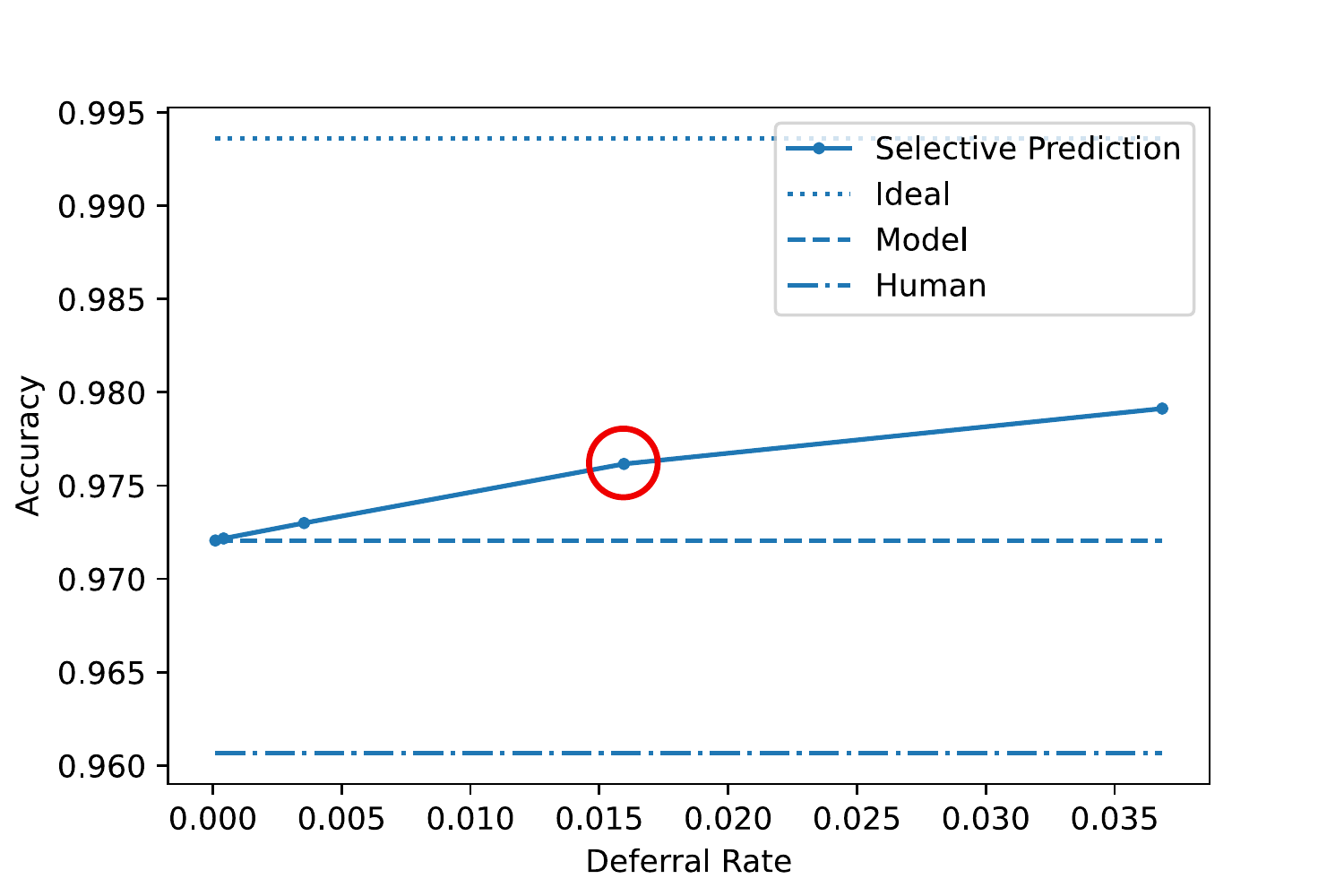}
    \caption{Tradeoff of expected accuracy and deferral rate. More deferral improves performance, but we still gain performance even when deferring less at the circled point.}
    \label{fig:tradeoff}
\end{figure}

\paragraph{Choosing a Deferral Model:} We now describe how we choose a deferral model on the \serengeti~dataset. Our goal is to find a model such that the accuracy of the $\text{sp}$ classifier is higher than the human $h$ or the AI ensemble $m$. 
A large fraction of the camera trap images are false positives, where the camera trap was triggered by a stimulus that was not an actual animal. In order to create a balanced dataset for tuning the deferral model, we subsample these empty images with no animal present. We implement a penalty for deferral for the $\text{sp}$ classifier, which leads to varied performance at different deferral rartes, 
as seen in Fig. \ref{fig:tradeoff}. 
In Fig. \ref{fig:tradeoff}, individual human accuracy (as opposed to consensus accuracy, which is 1.0) is 0.961, and AI model accuracy is 0.972 (based on choosing one operating point to turn the continuous model scores into binary predictions that maximizes accuracy of the AI-only classifier). We also include the ideal performance if we had a perfect oracle to decide, for each image, whether to defer to a human or rely on the model (given historical human labels). This perfect selective prediction would achieve an accuracy of 0.994. 

Given these results, we choose a deferral rate of $1\%$ as an acceptable level of withholding (see Appendix for details) that still improves expected accuracy by a significant margin relative to AI-only or human-only classifiers.

\section{Experiment Design}\label{sec:exp}

\begin{figure}
    \centering
    \includegraphics[width=0.45\textwidth]{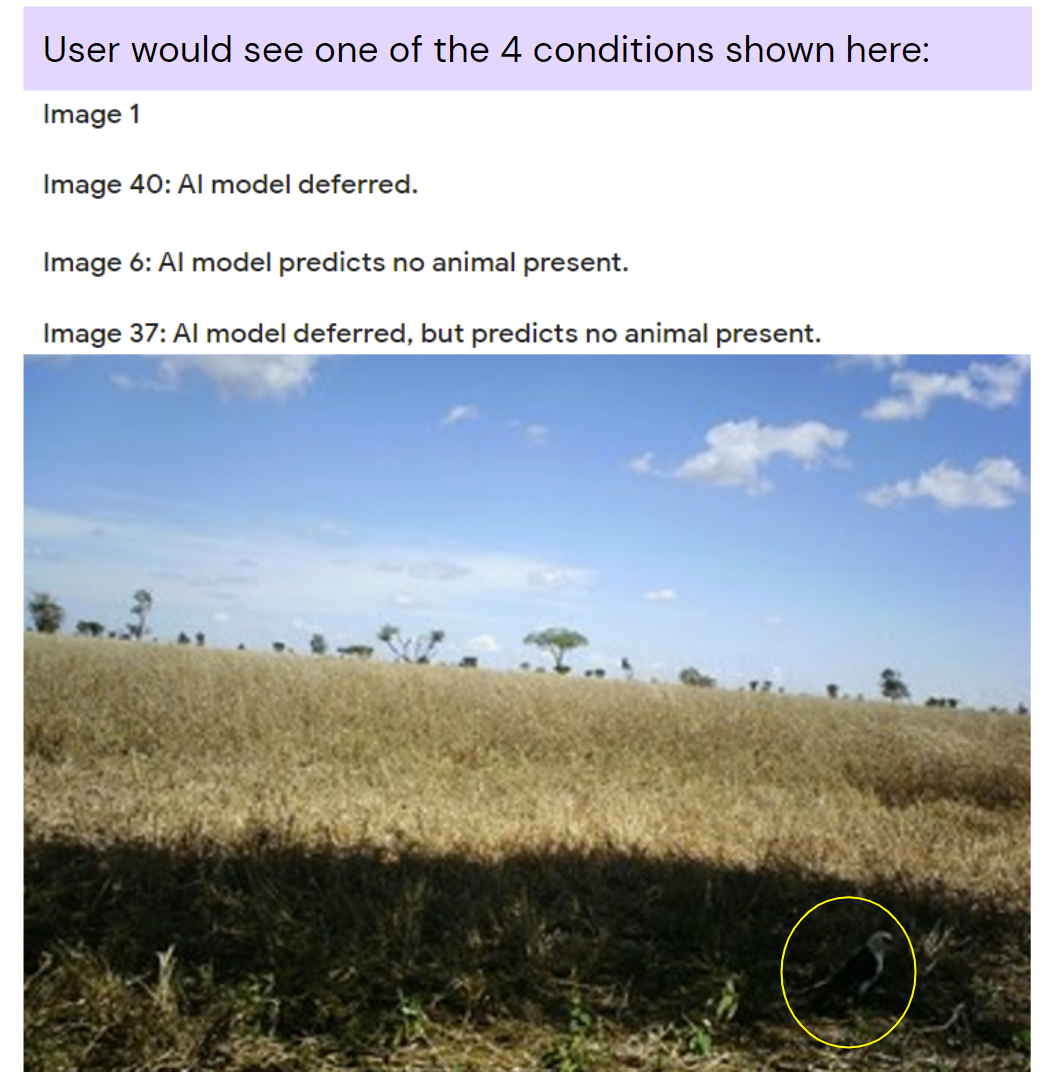}
     \includegraphics[width=0.45\textwidth]{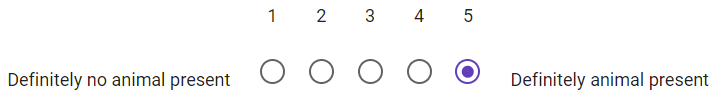}
    \caption{ The four possible \msg~conditions in our experiment, along with a challenging example image. The animal is circled in the image here for the reader's convenience, but in our experiment the circle was not present.}
    \label{fig:conditions}
\end{figure}

We hypothesize that the accuracy of a human decision maker in a human-AI team is affected by the \msg~in the last step of our workflow (Fig. \ref{fig:workflow}). We specifically consider presenting  information about the AI's \textbf{\predmain} and \textbf{\defermain}. The AI \predmain~refers to the class returned by the AI model, e.g., animal or no animal. The \defermain~refers to the result of selective prediction, in which we threshold the model score to determine whether to ask a human to review an image (defer), or rely on the AI prediction. 
We therefore design a human participant experiment with all possible combinations of these two details: 1) Neither message (\nomsg), 2) Deferral status only (\deferonly), 3) Prediction only (\predonly), and 4) Both messages (\both), as shown in Fig. \ref{fig:conditions}. 
We create a survey to host this experiment, consisting of the following sections: 1) Information and consent, 2) Camera trap training and explanation, in which participants are introduced to camera trap images and given an example with an explanation for the best label, 3) Adding AI assistance, in which participants are told about adding an AI model and deferral to assist in the task of sorting camera trap images, 4) AI assist practice and explanations, which consists of 10 examples (9 correct, 1 incorrect) drawn from the \serengeti~validation set with AI model assistance, 5) Post training questions asking participants to describe the model strengths and weaknesses, 6) Labeling, and 7) Post dataset questions. Several of these design choices align with the guidelines from \citet{amershi2019guidelines}, including describing the AI performance, and providing examples and explanations.

In the labeling section, we display 80 model-deferred images (like Fig. \ref{fig:conditions}) under the four \msg~conditions (yielding 20 images per condition, which we believe balances a reasonable number of examples with a manageable amount of participants’ time). Each includes a request for labels of animals present or not, with a Likert scale as in Fig. \ref{fig:conditions}.
The images are randomly allocated across the four communication conditions. \textbf{We did not inform participants that all images were deferred, we only relied on the different \msg~conditions.} Each participant judged images across all four conditions, and no single image was presented more than once to the same participant. The set of 80 images are sampled so as to  balance the number of true positive, false positive, true negative, and false negative model classifications, and therefore additionally ensure a balance in the number of images across classes. To ensure that there are no effects due to the specific order or allocation to a condition of each image, four separately seeded random allocations are carried out and each participant is randomly assigned to one of them. To test if the effects of the experimental conditions on accuracy exceed variation expected by chance, the data are analysed in a 2x2x2 within-subject repeated measures ANOVA with the factors ``\defermain'' (with the levels: ``shown'' and ``not shown''), ``\predmain'' (with the levels: ``shown'' and ``not shown'') and ``model accuracy'' (with the levels: ``model correct'' and ``model incorrect''). We received approval from an internal ethics review board, and then 
 recruited 198 participants from Prolific to take part in the experiment. Responses from all 198 participants were included in the ANOVA. Aggregated data are available at \url{https://github.com/deepmind/HAI_selective_prediction/}.

\begin{figure}
    \centering
    \includegraphics[width=0.35\textwidth]{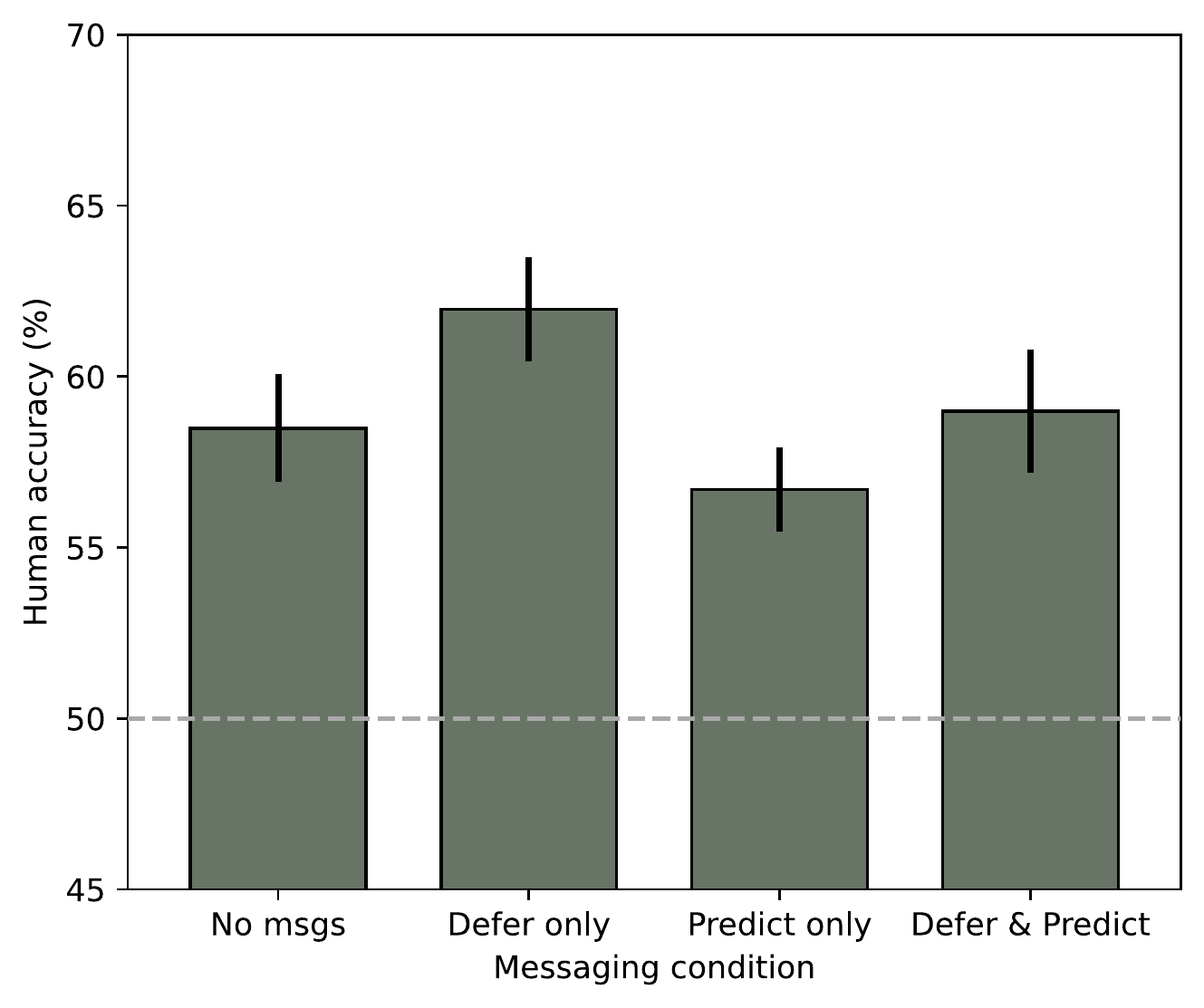}
    \caption{ 
    Accuracy of human participants on deferred images, across different \msg~conditions. Each bar shows the participant classification accuracies across the entire dataset, errorbars show 95\% confidence intervals on the mean. Participants' responses are more accurate when the images are accompanied by the context that the images are deferred (\deferonly~and \both~vs. \nomsg~and \predonly). Showing the model's prediction of the label has a negative effect on accuracy. The horizontal dashed grey line indicates chance performance (50\%).}
    \label{fig:human_maineffect}
\end{figure}

\section{Experiment Results and Analysis}

\subsection{\msg~Affects Human Accuracy}
As can be seen in Fig. \ref{fig:human_maineffect}, 
the information provided to human participants about the model affected participants' accuracy. The human-AI communication method that yields the highest human performance is \deferonly. Accuracy in this condition is significantly greater than either humans classifying images by themselves without any information about the model (mean of \deferonly: 61.9\%, mean of \nomsg: 58.4\%, $p<0.001$), or the model operating alone (mean of \deferonly: 61.9\%, mean of model alone: 50\%, $p<0.001$). Model performance is 50\% since the images presented to humans are subsampled from the set of model-deferred images. 
Furthermore, across all responses, participants are significantly more accurate when the \defermain~is shown (conditions \deferonly~and \both) than when it is not (conditions \predonly~and \nomsg) (mean of \defermain: 60.4\%, mean of no \defermain: 57.4\%, $p<0.001$). 
We believe this effect of \defermain~may be driven by participants inferring that the images are likely to be quite difficult, and therefore concentrating harder. By contrast, participants are significantly less accurate when the \predmain~is shown (mean of \predmain: 57.8\%, mean of no \predmain: 60.2\%, $p=0.003$). 
Overall, therefore, we find that providing the \defermain, while avoiding the provision of model predictions, leads to the highest accuracy in human decision-making in this context.

\subsection{Model Predictions Influence Human Decisions}
\begin{figure}
    \centering
    \includegraphics[width=0.35\textwidth]{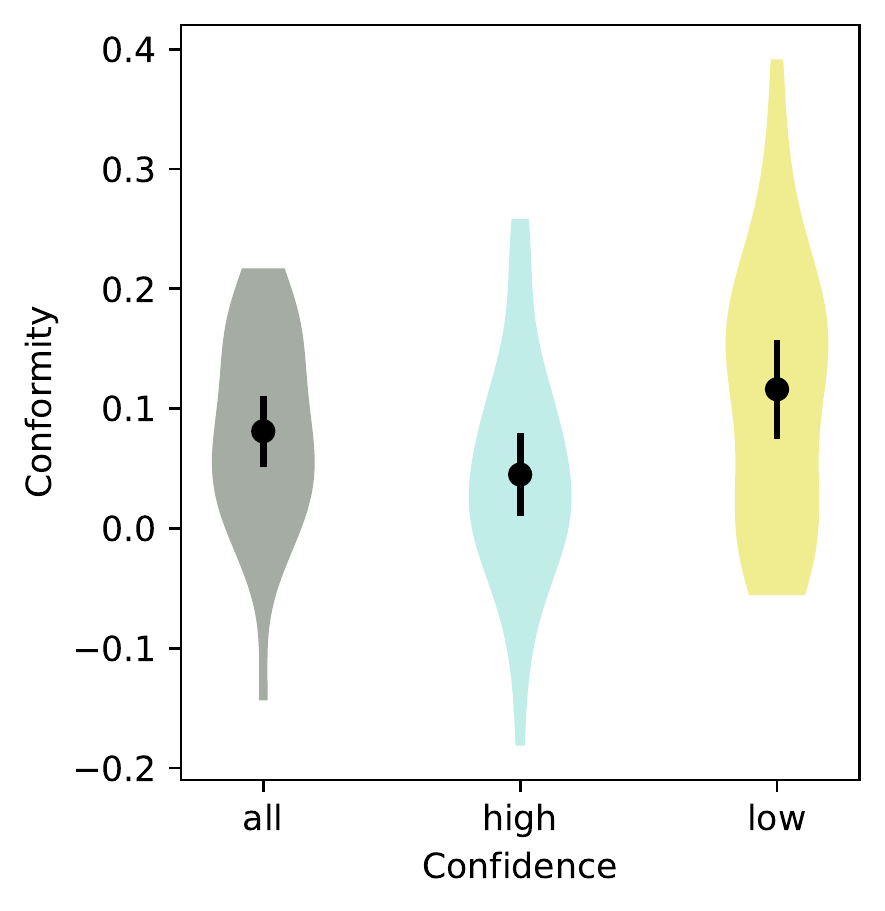}
    \caption{
    Conformity is the relative increase in agreement when the model predictions are present. For each image, we show how much the agreement between the set of human predictions and model prediction changes from the \nomsg~to the \predonly~conditions.  Conformity is significantly higher when humans are less confident, indicating that humans are influenced by model predictions more when they are less certain about their judgement.}
    \label{fig:human_conformity}
\end{figure}

While the preceding results clearly suggest that participants use the model predictions at least some of the time, the evidence is indirect. We therefore conduct a more direct analysis targeting this question. Specifically, for each image, we compute the proportion of raters who agree with the model prediction under 1) \nomsg~ and 2) \predonly. As each image is presented under both conditions, we subtract these two scores to assess how much humans increase their agreement with the model when the model prediction is present. We refer to this as the ``conformity'' score, and plot it in Fig. \ref{fig:human_conformity}. Across the set of images, we find an average conformity score of 0.08 (i.e., raters are influenced on 8\% of trials), which is significantly greater than zero (
$p< 0.0001$), demonstrating that raters use the model information when present. We further test a hypothesis suggested by \citet{bahrami2010optimally, BOORMAN20131558}, that people are more likely to use model information when they are less confident in their own decisions (as measured by Likert ratings in this case). We separately compute the conformity score for low and high confidence human decisions, and find that, as expected, conformity is higher when rater confidence is low (mean of low confidence: 0.116, mean of high confidence: 0.045, 
$p = 0.014$). 

\begin{figure}
    \centering
    \includegraphics[width=0.48\textwidth]{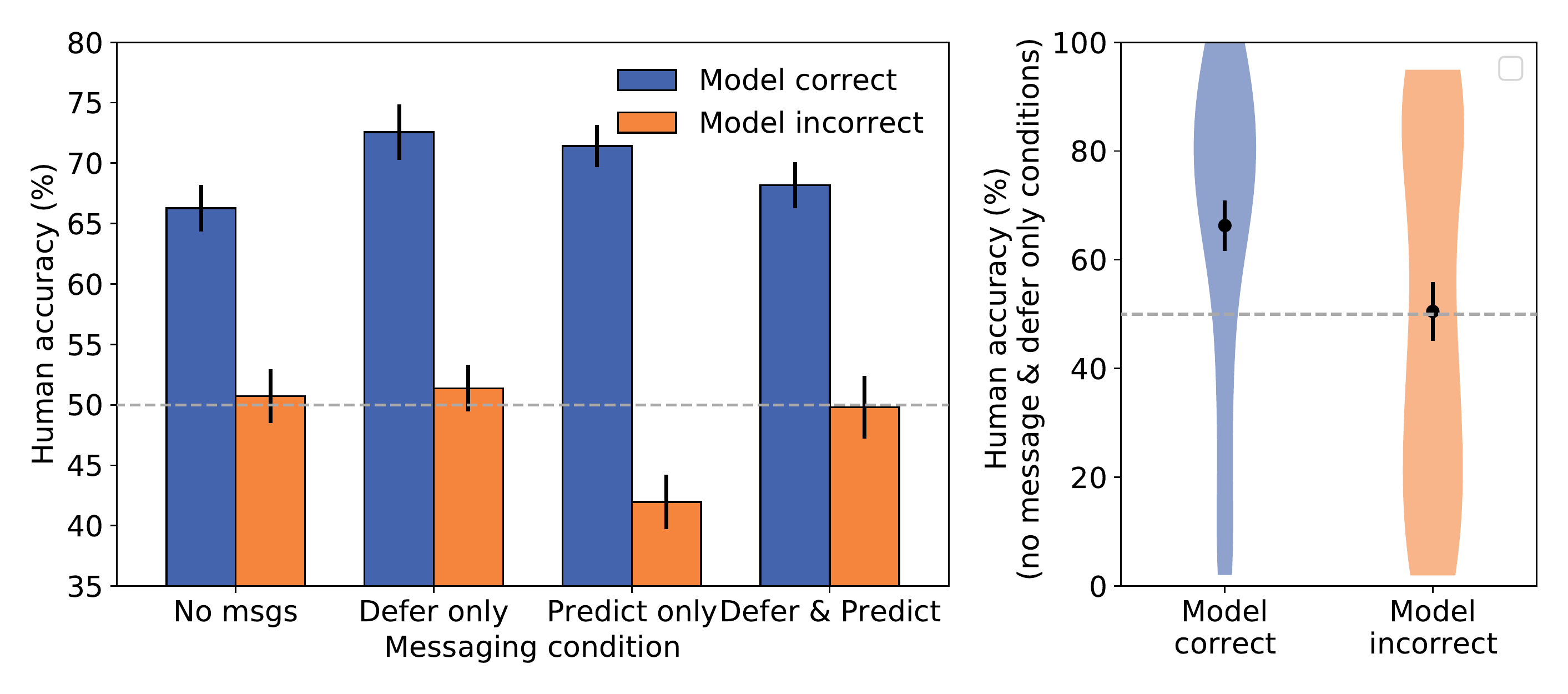}
    \caption{
    Accuracy of human participants on deferred images, split by whether the model correctly or incorrectly classified the associated images. (Left) Results split by whether the model correctly classified the image. Notably, images where the model is incorrect has lower participants' accuracy across all conditions,  generally at chance (dashed grey line). Crucially, participants are significantly below chance in the \predonly~condition, for which the model prediction misleads humans.  (Right) Human accuracy for each image in the \nomsg~and  \deferonly~conditions, split by whether the model labels the image correctly. Participants' accuracy is significantly reduced (to around chance) for the subset of images which the model fails to label correctly. Together, both show a congruence between what images the model and humans find difficult to label correctly.} 
    \label{fig:human_correctincorrect}
\end{figure}

\subsection{Model Accuracy Affects Human Accuracy}
While the preceding analysis demonstrates that people do indeed use the model predictions, this in itself does not explain why we find a decrease in human accuracy when model predictions are available. This suggests a potential bias, such that people tend to use the model information more when it is actually incorrect. To explore this possibility, we focus on the effect of the provided \msg~only on the images that the model classifies incorrectly (Fig. \ref{fig:human_correctincorrect}). We observe that participants' accuracy in the \predonly~condition is significantly reduced. While participants perform at chance in the other conditions, participants perform significantly below chance in the \predonly~condition (mean of other conditions: 50.6\%, mean of \predonly: 41.9\%, 
$p<0.001$). In this condition, the label provided by the model is most salient, and critically it provides wrong information. Participants appear to integrate this information as they perform $8.7\%$ worse in the \predonly~condition compared to the \nomsg~condition, on the images that are misclassified by the model. In contrast, on images that are correctly classified by the model, and therefore the model can provide a correct \predmain~message, participants gain $5.1\%$ of accuracy between the \nomsg~and \predonly~conditions. This asymmetry, that incorrect model \predmain~messages are integrated more by the participants than correct model \predmain~messages, appears to lie at the heart of why we observe an overall negative effect of the \predmain~message.
This pattern is consistent with the observation that participants and the model tend to err on the same images, as shown in the violin plots of Fig. \ref{fig:human_correctincorrect}. Specifically, participants are correct $66.3\%$ of the time on images shown in the \nomsg~and \deferonly~conditions and that the model correctly classifies, but only $50.5\%$ on images in these same conditions that the model classifies incorrectly (
$p<0.05$). Additionally, there is a positive correlation between average human Likert scores in the \nomsg~and \deferonly~conditions and model scores on the same images, suggesting that humans and models learn similar semantic task dimensions (Pearson's $r=0.27$, $p=0.021$). 


\subsection{Asymmetry in Human-AI Agreement}
Our results demonstrate a differential change in human accuracy in the presence of model information, based on whether the model is correct or incorrect. 
When we subdivide the data based on whether the model is correct or incorrect, we see that human accuracy tends to be lower when the model is incorrect (Fig. \ref{fig:human_correctincorrect}, left). 
Additionally,  exposing humans to incorrect predictions of an AI makes them even more likely to be incorrect. 
We confirm that human accuracy tends to be lower when the model is incorrect by directly computing the image-wise agreement between human and model ratings on images where the model is correct vs. incorrect (Fig. \ref{fig:human_correctincorrect}, right). This analysis is performed only on the trials under the \nomsg~condition, as we want to investigate independent agreement. As expected, we find that agreement is significantly higher for the correctly labelled images (mean agreement of model correct: 69.6\%, mean of model incorrect: 44.9\%, 
$p = 0.007$). 
This difference in prior agreement over images has potentially important repercussions when introducing model predictions. Specifically, as the subset of images where the model is correct already has high prior agreement with the human raters, there is much less potential for the model to influence human judgments, and hence increase accuracy. By contrast, because the images that are misclassified by the model have a lower prior agreement level with human raters, there is greater potential for model influence, which in this case will decrease accuracy. This asymmetry in potential model influence between correct and incorrect trials is likely to account for the overall drop in human accuracy we see when model predictions are provided to human raters. This is an important result, as it demonstrates that the specific pattern of covariance between model and human decisions can lead to significant downstream effects on joint decision-making when humans have access to the model predictions.

\section{Discussion}
To summarize, we find that there are significant effects in the way deferred images are presented to humans in a selective prediction workflow. In particular, in this context, presenting \defermain~is helpful, while presenting the uncertain \predmain, even when accompanied by a \defermain, can be harmful, especially when the model is incorrect. From this, it is clear that performance will not necessarily be what is estimated from historical human labels.  Together, these findings illustrate the importance of considering the human-AI team while designing selective prediction systems, as the \msg~can have a significant impact on performance.

We believe these are important findings to direct future research, and have several suggestions for open questions to explore. First, this experiment could be expanded. While we chose to leave this static, it would be helpful to determine the amount and type of training that is most useful for participants. We also chose to use two categories for \defermain, either defer or not defer. However, it is possible that finer-grained information about model uncertainty could be helpful \cite{bhatt2021uncertainty}. We additionally focused on understanding why \predmain~hurt, but encourage collecting further information, such as timing, to better understand why \defermain~helped. This may help inform further research into designing selective prediction algorithms based on human-AI teams, for example by exploring bounded rationality for training improved  selective prediction models. 

There are also questions about generalizability of these results. These specific results (i.e., that \defermain~helps while \predmain~hurts performance) are not likely to be robust across datasets, different human-AI use scenarios (e.g., decision aids), or even participant expertise levels. For example, in this study, we asked two domain experts working with the \serengeti~dataset to go through the same survey provided to Prolific participants. We similarly find that each individual has different performance in the four conditions, and that the \defermain~leads to improved performance in both cases.   However, the interactions are slightly different. It is necessary to search for generalizable trends across these cases in future work. 

Finally, though we worked with humans in this study, it is extremely important to consider specific deployment challenges in these contexts, such as how selective prediction may change existing processes, e.g., in healthcare \cite{yang2019unremarkable}, or how to best modify the workflow and instructions in the case where there are multiple human experts we could rely on.  
%
%
In all cases, we stress the importance of remembering the human component of human-AI teams.

\section{Acknowledgments}
We would like to thank the Prolific participants, Exavery Kigosi from the Grumeti Fund, DeepMind HuBREC, and the reviewers for their helpful feedback.

\section{Ethics Statement}
We received approval for this work from DeepMind's ethics review board, HuBREC 21 008. We did not collect personally identifiable information, and we paid a flat rate of £7 for the task, which took about 20-30 minutes. 
We believe it's unlikely that any of the participants were exposed to inappropriate imagery during the study. However, the camera trap datasets are made up of animals, both adult and young, in the wild, and thus could potentially include scenes of predator-prey interactions that some raters may find distressing.  
We included warnings about the species, behaviors, and possible human presence in the information sheet. Furthermore, we ensured participants understood that they were under no obligation to consent to participate, and that if they declined to participate, there would be no negative repercussions.

\bibliography{aaai22}

\begin{thebibliography}{32}
\providecommand{\natexlab}[1]{#1}

\bibitem[{Amershi et~al.(2019)Amershi, Weld, Vorvoreanu, Fourney, Nushi,
  Collisson, Suh, Iqbal, Bennett, Inkpen et~al.}]{amershi2019guidelines}
Amershi, S.; Weld, D.; Vorvoreanu, M.; Fourney, A.; Nushi, B.; Collisson, P.;
  Suh, J.; Iqbal, S.; Bennett, P.~N.; Inkpen, K.; et~al. 2019.
\newblock Guidelines for human-AI interaction.
\newblock In \emph{Proceedings of the 2019 chi conference on human factors in
  computing systems}, 1--13.

\bibitem[{Bahrami et~al.(2010)Bahrami, Olsen, Latham, Roepstorff, Rees, and
  Frith}]{bahrami2010optimally}
Bahrami, B.; Olsen, K.; Latham, P.~E.; Roepstorff, A.; Rees, G.; and Frith,
  C.~D. 2010.
\newblock Optimally interacting minds.
\newblock \emph{Science}, 329(5995): 1081--1085.

\bibitem[{Bansal et~al.(2021)Bansal, Nushi, Kamar, Horvitz, and
  Weld}]{bansal2021most}
Bansal, G.; Nushi, B.; Kamar, E.; Horvitz, E.; and Weld, D.~S. 2021.
\newblock Is the Most Accurate AI the Best Teammate? Optimizing AI for
  Teamwork.
\newblock In \emph{Proceedings of the AAAI Conference on Artificial
  Intelligence}, volume~35, 11405--11414.

\bibitem[{Beery et~al.(2020)Beery, Wu, Rathod, Votel, and
  Huang}]{beery2020context}
Beery, S.; Wu, G.; Rathod, V.; Votel, R.; and Huang, J. 2020.
\newblock Context r-cnn: Long term temporal context for per-camera object
  detection.
\newblock In \emph{Proceedings of the IEEE/CVF Conference on Computer Vision
  and Pattern Recognition}, 13075--13085.

\bibitem[{Bhatt et~al.(2021)Bhatt, Antor{\'a}n, Zhang, Liao, Sattigeri,
  Fogliato, Melan{\c{c}}on, Krishnan, Stanley, Tickoo
  et~al.}]{bhatt2021uncertainty}
Bhatt, U.; Antor{\'a}n, J.; Zhang, Y.; Liao, Q.~V.; Sattigeri, P.; Fogliato,
  R.; Melan{\c{c}}on, G.; Krishnan, R.; Stanley, J.; Tickoo, O.; et~al. 2021.
\newblock Uncertainty as a form of transparency: Measuring, communicating, and
  using uncertainty.
\newblock In \emph{Proceedings of the 2021 AAAI/ACM Conference on AI, Ethics,
  and Society}, 401--413.

\bibitem[{Boorman et~al.(2013)Boorman, O’Doherty, Adolphs, and
  Rangel}]{BOORMAN20131558}
Boorman, E.; O’Doherty, J.; Adolphs, R.; and Rangel, A. 2013.
\newblock The Behavioral and Neural Mechanisms Underlying the Tracking of
  Expertise.
\newblock \emph{Neuron}, 80(6): 1558--1571.

\bibitem[{Bu{\c{c}}inca, Malaya, and Gajos(2021)}]{buccinca2021trust}
Bu{\c{c}}inca, Z.; Malaya, M.~B.; and Gajos, K.~Z. 2021.
\newblock To trust or to think: cognitive forcing functions can reduce
  overreliance on AI in AI-assisted decision-making.
\newblock \emph{Proceedings of the ACM on Human-Computer Interaction},
  5(CSCW1): 1--21.

\bibitem[{Chow(1970)}]{chow1970optimum}
Chow, C. 1970.
\newblock On optimum recognition error and reject tradeoff.
\newblock \emph{IEEE Transactions on information theory}, 16(1): 41--46.

\bibitem[{De-Arteaga, Fogliato, and Chouldechova(2020)}]{de2020case}
De-Arteaga, M.; Fogliato, R.; and Chouldechova, A. 2020.
\newblock A case for humans-in-the-loop: Decisions in the presence of erroneous
  algorithmic scores.
\newblock In \emph{Proceedings of the 2020 CHI Conference on Human Factors in
  Computing Systems}, 1--12.

\bibitem[{Dietvorst, Simmons, and Massey(2015)}]{dietvorst2015algorithm}
Dietvorst, B.~J.; Simmons, J.~P.; and Massey, C. 2015.
\newblock Algorithm aversion: People erroneously avoid algorithms after seeing
  them err.
\newblock \emph{Journal of Experimental Psychology: General}, 144(1): 114.

\bibitem[{Galesic and Garcia-Retamero(2010)}]{galesic2010statistical}
Galesic, M.; and Garcia-Retamero, R. 2010.
\newblock Statistical numeracy for health: a cross-cultural comparison with
  probabilistic national samples.
\newblock \emph{Archives of internal medicine}, 170(5): 462--468.

\bibitem[{Gaube et~al.(2021)Gaube, Suresh, Raue, Merritt, Berkowitz, Lermer,
  Coughlin, Guttag, Colak, and Ghassemi}]{gaube2021ai}
Gaube, S.; Suresh, H.; Raue, M.; Merritt, A.; Berkowitz, S.~J.; Lermer, E.;
  Coughlin, J.~F.; Guttag, J.~V.; Colak, E.; and Ghassemi, M. 2021.
\newblock Do as AI say: susceptibility in deployment of clinical decision-aids.
\newblock \emph{NPJ digital medicine}, 4(1): 1--8.

\bibitem[{Geifman and El-Yaniv(2017)}]{geifman2017selective}
Geifman, Y.; and El-Yaniv, R. 2017.
\newblock Selective classification for deep neural networks.
\newblock \emph{arXiv preprint arXiv:1705.08500}.

\bibitem[{Green and Chen(2019)}]{green2019principles}
Green, B.; and Chen, Y. 2019.
\newblock The principles and limits of algorithm-in-the-loop decision making.
\newblock \emph{Proceedings of the ACM on Human-Computer Interaction}, 3(CSCW):
  1--24.

\bibitem[{Grote(2021)}]{grote2021trustworthy}
Grote, T. 2021.
\newblock Trustworthy medical AI systems need to know when they don’t know.
\newblock \emph{Journal of Medical Ethics}, 47(5): 337--338.

\bibitem[{Keswani, Lease, and Kenthapadi(2021)}]{keswani2021towards}
Keswani, V.; Lease, M.; and Kenthapadi, K. 2021.
\newblock Towards Unbiased and Accurate Deferral to Multiple Experts.
\newblock \emph{arXiv preprint arXiv:2102.13004}.

\bibitem[{Kompa, Snoek, and Beam(2021)}]{kompa2021second}
Kompa, B.; Snoek, J.; and Beam, A.~L. 2021.
\newblock Second opinion needed: communicating uncertainty in medical machine
  learning.
\newblock \emph{NPJ Digital Medicine}, 4(1): 1--6.

\bibitem[{Kool and Botvinick(2018)}]{kool2018mentallabour}
Kool, W.; and Botvinick, M. 2018.
\newblock Mental labour.
\newblock \emph{Nature Human Behaviour}, 2(5995): 899--908.

\bibitem[{Logg, Minson, and Moore(2019)}]{logg2019algorithm}
Logg, J.~M.; Minson, J.~A.; and Moore, D.~A. 2019.
\newblock Algorithm appreciation: People prefer algorithmic to human judgment.
\newblock \emph{Organizational Behavior and Human Decision Processes}, 151:
  90--103.

\bibitem[{Madras, Pitassi, and Zemel(2017)}]{madras2017predict}
Madras, D.; Pitassi, T.; and Zemel, R. 2017.
\newblock Predict responsibly: improving fairness and accuracy by learning to
  defer.
\newblock \emph{arXiv preprint arXiv:1711.06664}.

\bibitem[{Mozannar and Sontag(2020)}]{mozannar2020consistent}
Mozannar, H.; and Sontag, D. 2020.
\newblock Consistent estimators for learning to defer to an expert.
\newblock In \emph{International Conference on Machine Learning}, 7076--7087.
  PMLR.

\bibitem[{Norouzzadeh et~al.(2021)Norouzzadeh, Morris, Beery, Joshi, Jojic, and
  Clune}]{norouzzadeh2021deep}
Norouzzadeh, M.~S.; Morris, D.; Beery, S.; Joshi, N.; Jojic, N.; and Clune, J.
  2021.
\newblock A deep active learning system for species identification and counting
  in camera trap images.
\newblock \emph{Methods in Ecology and Evolution}, 12(1): 150--161.

\bibitem[{Norouzzadeh et~al.(2018)Norouzzadeh, Nguyen, Kosmala, Swanson,
  Palmer, Packer, and Clune}]{NorouzzadehE5716}
Norouzzadeh, M.~S.; Nguyen, A.; Kosmala, M.; Swanson, A.; Palmer, M.~S.;
  Packer, C.; and Clune, J. 2018.
\newblock Automatically identifying, counting, and describing wild animals in
  camera-trap images with deep learning.
\newblock \emph{Proceedings of the National Academy of Sciences}, 115(25):
  E5716--E5725.

\bibitem[{Rastogi et~al.(2022)Rastogi, Zhang, Wei, Varshney, Dhurandhar, and
  Tomsett}]{rastogi2020deciding}
Rastogi, C.; Zhang, Y.; Wei, D.; Varshney, K.~R.; Dhurandhar, A.; and Tomsett,
  R. 2022.
\newblock Deciding Fast and Slow: The Role of Cognitive Biases in AI-assisted
  Decision-making.
\newblock In \emph{CSCW}.

\bibitem[{Swanson et~al.(2015)Swanson, Kosmala, Lintott, Simpson, Smith, and
  Packer}]{swanson2015snapshot}
Swanson, A.; Kosmala, M.; Lintott, C.; Simpson, R.; Smith, A.; and Packer, C.
  2015.
\newblock Snapshot Serengeti, high-frequency annotated camera trap images of 40
  mammalian species in an African savanna.
\newblock \emph{Scientific data}, 2(1): 1--14.

\bibitem[{Tabak et~al.(2019)Tabak, Norouzzadeh, Wolfson, Sweeney, VerCauteren,
  Snow, Halseth, Di~Salvo, Lewis, White et~al.}]{tabak2019machine}
Tabak, M.~A.; Norouzzadeh, M.~S.; Wolfson, D.~W.; Sweeney, S.~J.; VerCauteren,
  K.~C.; Snow, N.~P.; Halseth, J.~M.; Di~Salvo, P.~A.; Lewis, J.~S.; White,
  M.~D.; et~al. 2019.
\newblock Machine learning to classify animal species in camera trap images:
  Applications in ecology.
\newblock \emph{Methods in Ecology and Evolution}, 10(4): 585--590.

\bibitem[{Tversky and Kahneman(1981)}]{tversky1981framing}
Tversky, A.; and Kahneman, D. 1981.
\newblock The framing of decisions and the psychology of choice.
\newblock \emph{Science}, 211(4481): 453--458.

\bibitem[{Wiener and El-Yaniv(2013)}]{wiener2013theoretical}
Wiener, Y.; and El-Yaniv, R. 2013.
\newblock \emph{Theoretical foundations of selective prediction}.
\newblock Ph.D. thesis, Computer Science Department, Technion.

\bibitem[{Wilder, Horvitz, and Kamar(2020)}]{wilder2020learning}
Wilder, B.; Horvitz, E.; and Kamar, E. 2020.
\newblock Learning to complement humans.
\newblock \emph{arXiv preprint arXiv:2005.00582}.

\bibitem[{Willi et~al.(2019)Willi, Pitman, Cardoso, Locke, Swanson, Boyer,
  Veldthuis, and Fortson}]{willi2019identifying}
Willi, M.; Pitman, R.~T.; Cardoso, A.~W.; Locke, C.; Swanson, A.; Boyer, A.;
  Veldthuis, M.; and Fortson, L. 2019.
\newblock Identifying animal species in camera trap images using deep learning
  and citizen science.
\newblock \emph{Methods in Ecology and Evolution}, 10(1): 80--91.

\bibitem[{Yang, Steinfeld, and Zimmerman(2019)}]{yang2019unremarkable}
Yang, Q.; Steinfeld, A.; and Zimmerman, J. 2019.
\newblock Unremarkable ai: Fitting intelligent decision support into critical,
  clinical decision-making processes.
\newblock In \emph{Proceedings of the 2019 CHI Conference on Human Factors in
  Computing Systems}, 1--11.

\bibitem[{Zhang and Maloney(2012)}]{zhang2012ubiquitous}
Zhang, H.; and Maloney, L.~T. 2012.
\newblock Ubiquitous log odds: a common representation of probability and
  frequency distortion in perception, action, and cognition.
\newblock \emph{Frontiers in neuroscience}, 6: 1.

\end{thebibliography}

\clearpage
In this Appendix, we will provide additional details on the \serengeti~dataset, choosing the deferral model, and further details regarding the human experiment and results.

\section{\serengeti~Dataset}
\subsection{Data Details}
The \serengeti~dataset is made up of animals, both adult and young, in the wild.  Behaviors labeled in the dataset include standing, resting, moving, eating, and interacting. In addition, there may be some images with humans captured accidentally. We removed images that had been previously labeled to include humans, but there is a possibility that there are humans that were missed by previous labelers.
For reference, the animal species captured include Grant’s gazelles, reedbuck, dik dik, zebra, porcupine, Thomson’s gazelles, spotted hyena, warthog, impala, elephant, giraffe, mongoose, buffalo, hartebeest, guinea fowl, wildebeest, leopard, ostrich, lion, kori bustard, other bird, bat eared fox, bushbuck, jackal, cheetah, eland, aardwolf, hippopotamus, striped hyena, aardvark, hare, baboon, vervet monkey, waterbuck, secretary bird, serval, topi, honey badger, rodents, wildcat, civet, genet, caracal, rhinoceros, reptiles, zorilla.

There are a variable number of human labels per image, as labels are collected until consensus is reached. All images have at least 5 labels, so we sample 5 randomly for each image. From the sampled data, we calculate the mean Cohen's kappa value to be 0.886, meaning very high agreement on a scale of -1 to 1. In fact, individual humans achieve 0.973 accuracy compared to consensus. While consensus is not guaranteed to be correct, \citet{swanson2015snapshot} show that on a gold standard dataset in which experts and many crowdsourced contributors label images, there was 96\% agreement. We note that because some images were labeled in groupings of three due to the capture pattern and assigned the same labels for all, there may be some instances where one of the images is blank as the animal moves out of frame. 
From a random sample of 400 images, only 3 were mislabeled in the training set as containing animal species when there did not appear to be any present. These were when, out of the three pictures taken, an animal moved out of the camera's field of view in (usually) the last image.

\subsection{Deferral Model Details}

Our deferral model's objective function maximizes accuracy. Specifically, we define this as a weighted combination of sensitivity, the accuracy based on ground truth positive examples, and specificity, the accuracy based on ground truth negative examples. The weights to get the standard measure of accuracy are typically the number of positive and negative examples, respectively, but we allow them to be tuned to achieve different tradeoffs if desired. In our case, we choose a model based on weighting sensitivity and specificity of the composite model equally in the objective function, at 0.5 each. 
Fig. 3 in the main paper is generated by modifying the penalty of withholding between -0.5 and -0.1, inclusive, by -0.1.

We choose the point which achieves a deferral rate of 0.01. At a deferral rate of 0.01, if one SD card containing 5k images is processed at a time, about 50 images are deferred to a human. With about 20 SD cards per month, this leads to about 1000 images for human review, compared to 109k. At about 5-10 seconds for the difficult images, and 1 second for the easy images, this means that we ask for a maximum of about 3 hours of human time, compared to 302 hours. 

Using this model, out of about 150k images in the \serengeti~test set, a total of 1297 images are deferred, with 603 images containing animals and 694 empty images. We find some degree of complementarity, in that model accuracy on non-deferred images is 0.978, but 0.577 on deferred images. Furthermore, we mostly defer on empty images, as humans tend to get images containing animals incorrect (in the 15990 cases out of about 150k in which they are incorrect). 

\section{Human Experiment Details}

We provide details of the ethical review in the ethics statement of the main paper. Further details are included below.

\subsection{Eligibility Criteria}
We required participants to be  consenting adults over 18 years old with intermediate-advanced English skills and good physical and mental health. Participants were requested from the UK and the US to increase the likelihood of intermediate-advanced English skills. Tasks were offered to workers via Prolific, an online automated system. Participants could abandon the task at any time. Domain experts were recruited by email.

\subsection{Survey}
We provide a link to one randomized version of the survey:  \url{https://bit.ly/SPM-Survey-AAAI2022}. 



\subsection{Additional Results Details}


Finally, we summarize our statistical findings from the main paper in Table \ref{tab:summary}, and provide the full ANOVA in Table \ref{tab:anova}. 

\begin{table*}[t]
\centering
\begin{tabular}{|c|c|c|} \hline
	\textbf{Comparison} & \textbf{Mean Values} & \textbf{Statistics (all significant)} \\\hline
	\deferonly~vs. \nomsg & 61.9, 58.4 & $t(197) = 3.8, p < 0.001$ \\\hline
	\deferonly~vs. model alone (chance)  & 61.9, 50 &  $t(197) = 15.4, p < 0.001$\\\hline
	\defermain~vs. no \defermain  & 60.4, 57.4 & $F(1, 197) = 18.9, p < 0.001$ \\\hline
	\predmain~vs. no \predmain  & 57.8, 60.2 & $F(1, 197) = 9.07, p = 0.003$ \\\hline
	Conformity score $> 0$  & 0.08, 0 & $t = 5.22, p < 0.0001$ \\\hline
	Conformity score for low $>$ high rater confidence & 0.116, 0.045 & $t = 2.54, p = 0.014$   \\\hline
	\predonly~vs. other conditions (when model incorrect) &  41.9, 50.6 & $t(197) = 23.9, p<0.001$ \\\hline
	Model correct vs incorrect (images in \nomsg~and \deferonly)  & 66.3, 50.5 & $F(1, 39)=2.19, p<0.05$ \\\hline
	Agreement b/w human and model ratings where model correct vs. incorrect  & 69.6, 44.9  & $t = 2.82, p = 0.007$ \\\hline
\end{tabular}
\caption{Summary of statistical results from human subject experiments.}
\label{tab:summary}
\end{table*}

\begin{table*}[]
\centering
\begin{tabular}{|c|c|c|c|c|}\hline
	&	\textbf{F Value} &	\textbf{Num DF}	& \textbf{Den DF} &	\textbf{Pr $>$ F}\\\hline
	\defermain &	18.9446&	1.0000&	197.0000&	0.0000 \\\hline
	\predmain &	9.0790&	1.0000&	197.0000&	0.0029 \\\hline
	model correct&	859.1269&	1.0000&	197.0000&	0.0000 \\\hline
	\defermain:\predmain&	0.7850&	1.0000&	197.0000&	0.3767 \\\hline
	\defermain:model correct&	2.3091&	1.0000&	197.0000&	0.1302 \\\hline
	\predmain:model correct&	12.5286&	1.0000&	197.0000&	0.0005 \\\hline
	\defermain:\predmain:model correct&	54.7870&	1.0000&	197.0000&	0.0000\\\hline
\end{tabular}
\caption{Full 2x2x2 ANOVA Results.}
\label{tab:anova}
\end{table*}

\end{document}